\documentclass[10pt,letterpaper]{article}
\usepackage[top=0.85in,left=2.75in,footskip=0.75in]{geometry}

\usepackage{amsmath,amssymb}
\usepackage{changepage}
\usepackage{textcomp,marvosym}
\usepackage{cite}
\usepackage{nameref,hyperref}
\usepackage[right]{lineno}
\usepackage[nopatch=eqnum]{microtype}
\DisableLigatures[f]{encoding = *, family = * }
\usepackage[table]{xcolor}
\usepackage{tikz}
\usetikzlibrary{positioning, arrows.meta, shapes.geometric, fit, backgrounds, calc}
\usepackage{array}

\newcolumntype{+}{!{\vrule width 2pt}}
\newlength\savedwidth

\newcommand\thickhline{\noalign{\global\savedwidth\arrayrulewidth\global\arrayrulewidth 2pt}%
\hline
\noalign{\global\arrayrulewidth\savedwidth}}

\raggedright
\usepackage[aboveskip=1pt,labelfont=bf,labelsep=period,justification=raggedright,singlelinecheck=off]{caption}

\makeatletter
\renewcommand{\@biblabel}[1]{\quad#1.}
\makeatother

\usepackage{lastpage,fancyhdr,graphicx}
\usepackage{epstopdf}
\fancyheadoffset[L]{2.25in}
\fancyfootoffset[L]{2.25in}
\begin{document}
\vspace*{0.2in}

\begin{flushleft}
{\Large
\textbf\newline{Multitask Multimodal Fusion with Tabular Foundation Models for Peak and Durability Prediction of Pertussis Booster Response}
}
\newline
\\
Divya Sitani
\\
\bigskip
Berlin, Germany
\\
\bigskip
\end{flushleft}


\section*{Abstract}
Pertussis booster vaccination produces immune responses that vary widely across individuals in both peak magnitude and long-term durability. These two phases are governed by partly distinct biological compartments: peak reflects acute B-cell activation and antibody secretion, while durability reflects the establishment of long-term humoral memory~\cite{Amanna2007,Gillard2024}. Yet most computational models target only one, missing the full boost-and-wane trajectory. Jointly predicting both is non-trivial because the two endpoints are biologically dissociated rather than redundant; samples are small, modalities are heterogeneous with structured missingness, and the two tasks rely on different measurement windows.

We propose a multi-task contrastive multimodal fusion architecture combining frozen TabPFN-v2 per-modality encoders, a dual-label supervised contrastive loss that treats two subjects as a positive pair if they agree on the Task 1 label or the Task 2 label, modality dropout calibrated to empirical missingness, and missingness-masked attention fusion. Applied to the curated subset of CMI-PB data~\cite{CMIPB} ($n = 158$ subjects, four modalities, $44.9\%$ with at least one modality missing; Spearman $r = -0.58$ between peak and durability, $n = 96$), the model achieves test area under the receiver operating characteristic curve (AUROC) $0.797$ (95\% confidence interval [CI] $[0.621, 0.948]$) for peak response and $0.755$ (95\% CI $[0.519, 0.945]$) for durability, with both significant under joint
label permutation ($N = 1000$; $p = 0.002$ and $p = 0.045$). Across
logistic regression, XGBoost, and MLP baselines on raw features and
on TabPFN embeddings, the proposed model is the only one whose
95\% CIs lie above chance on both tasks simultaneously. Per-modality
contribution analyses recover task-specific modality contributions consistent with
the underlying immunology: peak prediction is carried by cytokine signatures, while durability is carried by baseline antibody features.

\clearpage
\newgeometry{top=0.85in,left=1in,right=1in,footskip=0.75in}

\section*{Introduction}

Pertussis, or whooping cough, has re-emerged as a significant public
health concern despite high vaccination coverage in many
countries~\cite{Mooi2014,DomenechDeCelles2024}. The introduction of
acellular pertussis (aP) vaccines in the 1990s, motivated by the
reactogenicity of earlier whole-cell pertussis (wP) formulations,
reduced adverse events but introduced a new challenge: aP-induced
immunity wanes more rapidly than wP-induced immunity, and aP vaccination
fails to prevent colonization and transmission of \textit{Bordetella
pertussis}~\cite{Warfel2014}. Epidemiological studies have documented
disproportionately high pertussis rates among adolescents and young
adults who received exclusively aP vaccines in
childhood~\cite{Sheridan2014}, and immunological investigations have revealed qualitative differences in the immune memory established by the two vaccine types. Whole-cell vaccines induce a Th1/Th17-polarized response and respiratory tissue-resident memory T cells that can persist for decades~\cite{McCarthy2024, daSilvaAntunes2018}, consistent with murine studies showing wP-induced T\textsubscript{RM}-mediated mucosal protection~\cite{Wilk2019}, whereas acellular vaccines elicit a Th2-biased, antibody-focused response that is effective against disease but less durable and less capable of preventing nasal colonization in animal models~\cite{daSilvaAntunes2021,Warfel2014}. Understanding what
determines the strength and longevity of booster responses, and how
these differ between priming regimens, is central to improving pertussis
vaccination strategies.

A critical but underappreciated aspect of vaccine response assessment
is that response magnitude and response durability are not the same
quantity measured at two timepoints; they are biologically distinct
processes with partially opposing dynamics. Subjects who mount the strongest peak antibody responses after boosting often exhibit the most rapid subsequent decline: the ``boost-and-wane'' trade-off, reflecting partly distinct biological programs governing acute and long-term humoral responses~\cite{Amanna2007,Gillard2024}. This dissociation has practical
consequences: a subject classified as a strong responder at day~14 may
be a poor retainer at day~120, and vice versa. Yet most computational
models of vaccine response treat prediction as a single-task problem,
typically targeting peak antibody titer or fold change alone. This
conflation obscures the distinct immunological programs underlying each
phase and misses the clinically relevant question of whether a given
individual will sustain protective antibody levels over months to years.
Jointly modeling both response phases, while respecting their biological
independence, requires a multi-task learning framework that can leverage
shared structure without forcing the two outcomes onto a single
predictive axis.

The CMI-PB (Computational Models of Immunity: Pertussis Boost)
project~\cite{CMIPB} provides a uniquely suited resource for this
challenge, building on the systems vaccinology paradigm pioneered by predictive analyses of yellow fever and influenza responses~\cite{Querec2009}. CMI-PB has profiled pertussis booster vaccine responses
across four annual cohorts (2020-2023) using four data modalities:
antibody titers, cytokine concentrations, cell frequencies, and gene
expression. The dataset has been the basis for community prediction
challenges, in which participants attempt to predict immune response
outcomes from pre-vaccination measurements~\cite{CMIPB}. However, prior
modeling efforts have largely been single-task (predicting peak antibody
response only), single-modality (using one data type at a time or simple
concatenation), and have not addressed the substantial structured
missingness inherent in multi-omic clinical data, where 44.9\% of
subjects in the Task~1 subset lack one or more complete modalities,
with missingness rates that can be up to 38.6\% (cytokine).

Tabular foundation models such as TabPFN~\cite{TabPFN} have recently
demonstrated strong performance on small tabular classification problems
through in-context learning, but their application to multimodal health
data with heterogeneous missingness remains largely unexplored. The
combination of small sample size ($n = 158$ subjects with the primary
label, $n = 96$ of these additionally with the durability label),
four heterogeneous
modalities, structured missingness, and two biologically anti-correlated
prediction tasks makes pertussis booster response prediction a
compelling testbed for multimodal learning methods designed for the
small-data regime.

Here, we propose a multi-task contrastive multimodal fusion architecture
for jointly predicting peak and durability of pertussis booster antibody
response. The model encodes each data modality independently using
frozen TabPFN-v2 embeddings~\cite{TabPFNv2}, projects them to a shared
representation space via independent projection heads with $l_2$
normalization, and aligns cross-modal embeddings using a dual-label
supervised contrastive loss that treats subjects as positive pairs if
they share the same label on either task. Modality dropout during
training, calibrated to the empirical missingness rate, ensures
robustness to arbitrary patterns of missing modalities at inference
time. Attention-weighted fusion with missingness-aware renormalization
combines available modality representations, which then pass through a
shared MLP before branching into task-specific classification heads.

Applied to CMI-PB data, the model achieves test AUROC $0.797$ for peak
response and $0.755$ for durability, with bootstrap 95\% confidence
intervals quantifying uncertainty on both estimates, and joint
label-permutation tests ($N = 1000$) rejecting the null of no
feature-label relationship at $p = 0.002$ (Task~1) and $p = 0.045$
(Task~2). Beyond predictive performance, per-modality contribution
analyses reveal that different modalities matter for different tasks(Fig~\ref{fig:modality_contribution}):
for peak prediction, cytokine signatures both dominate as the standalone predictor and are the only modality whose removal materially harms the ensemble; for durability, antibody features carry most of the information. We further
observe an asymmetric peak-durability dissociation (Fig~\ref{fig:motivation}): the
``low-peak/low-durability'' quadrant is essentially empty, with subjects
instead sorting into a ``big-and-fading'' or ``modest-and-durable''
response strategy. Architectural ablations (Table~\ref{tab:ablation}) show that both the
dual-label contrastive loss and modality dropout contribute substantial,
non-redundant performance gains, with a non-trivial interaction between
them.

\section*{Materials and methods}

\subsection*{Dataset}

\begin{figure*}[tbp]
\centering
\includegraphics[width=0.98\textwidth]{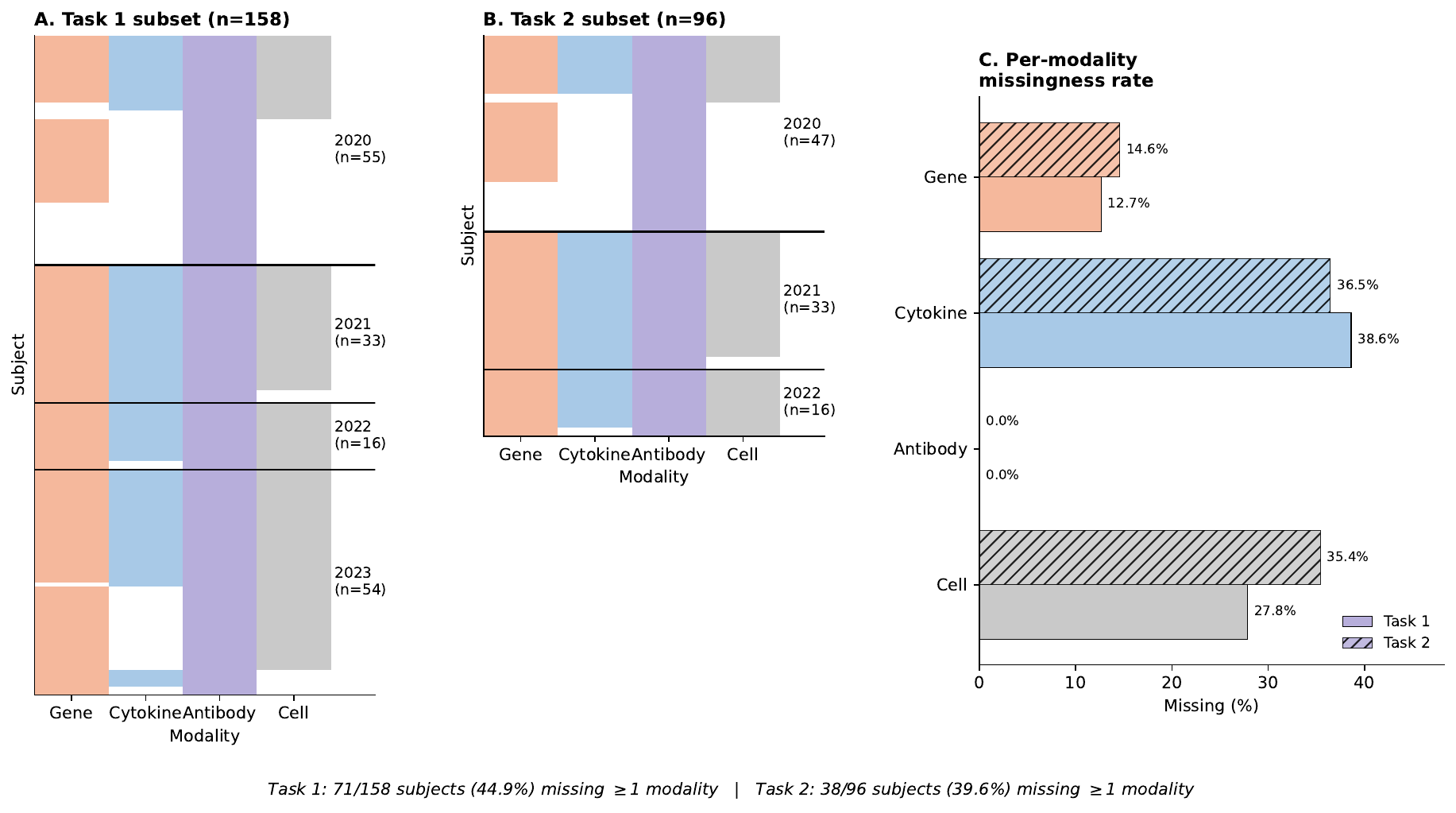}
\caption{\textbf{Cohort $\times$ modality missingness pattern in
CMI-PB.} \textbf{(A)}~Subject level modality availability for the
Task~1 (peak response) cohort, organised by annual cohort
(2020-2023). Each row is one subject; each column is one of the four
data modalities. Filled cells indicate the modality is present;
white cells indicate it is missing. Within each cohort, subjects are
sorted by missingness pattern (subjects with more modalities at the
top). Horizontal black lines separate cohorts. \textbf{(B)}~The same
visualization restricted to the Task~2 (durability) modelling subset:
subjects with both a Task~1 and a Task~2 label ($n = 96$). Subjects
with a Task~2 label but no Task~1 label ($n = 4$) are not modeled and
are excluded here. The 2023 cohort is absent because antibody
measurements at day 120 were not collected for that year. \textbf{(C)}~Per-modality
missingness rate within each task subset. The pattern reveals that missingness is structural and cohort driven, rather than reflecting random subject dropout: certain modalities were systematically not collected in certain cohort years. This motivates an architecture that handles arbitrary
modality subsets at both training and inference time.}
\label{fig:missingness}
\end{figure*}

We use the CMI-PB (Computational Models of Immunity: Pertussis Boost)
dataset~\cite{CMIPB}, a longitudinal multi-omic resource designed for benchmarking computational models of pertussis booster vaccination response. The
dataset comprises immune profiling of subjects receiving a pertussis
booster vaccination (Tdap), and we use four data modalities from CMI-PB measured across
multiple timepoints: plasma antibody titers (IgG subtype responses to
pertussis and control antigens), plasma cytokine concentrations (Olink
NPX), peripheral blood mononuclear cell (PBMC) frequencies by flow
cytometry, and PBMC gene expression (bulk RNA-seq). All four modalities were batch corrected by the CMI-PB consortium using the standardized pipeline described at \url{https://github.com/CMI-PB/cmi-pb-3rd-public-challenge-data-prep}: KNN imputation of missing values~\cite{Troyanskaya2001}, median normalization per dataset, and ComBat based batch correction~\cite{ComBat2007} of annual cohort-level batch effects~\cite{CMIPB}. Subjects span four annual
cohorts (2020--2023). From the CMI-PB resource, we retained subjects with at least one of the four modalities measured and an available label, yielding $n = 158$ for Task~1 (peak) and $n = 102$ for Task~2 (durability), with $n = 96$ subjects in common. The 2023 cohort lacks day 120 antibody measurements and is therefore automatically excluded from Task~2 via masking of missing labels. The modeled dataset now consists of the
$n = 158$ subjects with a Task~1 label and at least one modality;
$n = 4$ subjects with a Task~2 label but no Task~1 label are excluded
so that every modeled subject contributes to the primary task. Of the
$158$ modeled subjects, $n = 96$ additionally carry a Task~2 label and contribute to the durability head via masked loss (Fig.~\ref{fig:missingness}; see the \emph{Data split} section below).

Modality availability varies by cohort and assay. In the Task~1 subset ($n = 158$): antibody titers are nearly complete (0\% missing),
gene expression is also quite complete (12.7\% missing), while cell
frequency (27.8\%) and cytokine (38.6\%) show substantial missingness.
Only 55.1\% of Task~1 subjects have all four modalities; 44.9\% are
missing at least one (Fig~\ref{fig:missingness}). The Task~2 modeling subset ($n = 96$) shows a similar overall pattern, with cell, cytokine,
and gene missingness rates of 35.4\%, 36.5\%, and 14.6\% respectively;
39.6\% of Task~2 subjects are missing at least one modality. This
missingness is structured rather than random: it is driven primarily
by assay availability at the cohort level (certain modalities were not
collected in certain cohort years) rather than by subject-level
dropout. This motivates an architecture that handles arbitrary
modality subsets at both training and inference time, since
restricting to complete-data subjects would exclude entire cohorts. 

\subsection*{Prediction tasks}

We define two complementary binary classification tasks reflecting
biologically dissociated phases of the humoral response to pertussis
booster vaccination.

\subsubsection*{Task 1 (primary): peak response.}
The primary prediction target is the $\log_2$ fold change of IgG
anti-pertussis toxin (IgG-PT) antibody titer from day~0 to day~14:
\begin{eqnarray}
\label{eq:fc_peak}
\text{FC}_{\text{peak}} = \log_2\!\left(\frac{\text{IgG-PT}_{d14} + 1}{\text{IgG-PT}_{d0} + 1}\right)
\end{eqnarray}
A binary responder label is derived by median split (cutoff $1.254$).
By childhood vaccination history, wP-primed subjects are
disproportionately responders ($47$ responder / $32$ non-responder),
while aP-primed subjects are disproportionately non-responders
($31$ / $48$), consistent with the established wP advantage in peak
response magnitude. Pertussis toxin (PT) is the only antigen unique
to \textit{Bordetella pertussis}; anti-PT IgG is the standard
serological correlate of pertussis-specific immunity and the primary
endpoint in CMI-PB challenges~\cite{CMIPB}.

\subsubsection*{Task 2 (secondary): durability / retention}
The secondary prediction target captures antibody retention between the
near-peak timepoint (day~30) and long-term follow-up (day~120):
\begin{eqnarray}
\label{eq:fc_retention}
\text{FC}_{\text{retention}} = \log_2\!\left(\frac{\text{IgG-PT}_{d120} + 1}{\text{IgG-PT}_{d30} + 1}\right)
\end{eqnarray}
The denominator uses d30 rather than d0 to avoid a ceiling effect artifact in the shared baseline that would otherwise produce a spurious positive
correlation between peak and retention. The binary durability label is
derived by median split. A total of $n = 102$ subjects have complete
d30 and d120 measurements (2020-2022 cohorts); of these, $n = 96$
also carry a Task~1 label and constitute the modeled Task~2 cohort.
The remaining $n = 4$ subjects with Task~2 labels but no Task~1 label
are excluded. Median $\log_2$ retention is $-0.464$, corresponding to a typical
$\sim 28\%$ loss of d30 antibody level by d120. Missing Task~2
labels in the modeling subset ($n = 62$ subjects without a d120
measurement, primarily the 2023 cohort) are encoded as $-1$ and
handled by masked loss during training.

\subsubsection*{Peak-durability dissociation}
The two tasks capture anti-correlated biology: Spearman $r = -0.580$
($p = 3.97 \times 10^{-10}$, $n = 96$); Cohen's $\kappa = -0.520$
on the binary labels (Fig~\ref{fig:motivation}). A strong peak responder is substantially more likely to exhibit poor retention: the so-called ``boost-and-wane'' pattern~\cite{Gillard2024}. The
two tasks are therefore statistically dissociated rather than
redundant, motivating the multi-task design.

\subsection*{Input feature design}

Input features consist of baseline day~0 values and $\log_2$ fold
changes at non-label timepoints (Table~\ref{tab:design}). Day 1
features are included for cytokine and cell frequency, which capture
fast innate-response dynamics that peak within $24$-$48$ hours of
booster administration~\cite{Pulendran2014, Querec2009}. Antibody
and gene expression skip day~1: antibody titers do not change
appreciably one day after a booster, and the most informative
transcriptional signatures of vaccine response have been shown to
emerge at day~7~\cite{Nakaya2011, Li2014}.

For linear-scale modalities (antibody, cell):
$\text{lfc}(f, t) = \log_2((f_t + 1)/(f_0 + 1))$. For log-scale
modalities (cytokine NPX, gene expression):
$\text{lfc}(f, t) = f_t - f_0$.

\begin{table}[!ht]
\centering
\caption{{\bf Per-modality experimental design.} Label-defining
timepoints excluded from inputs. The PT family (IgG-PT and its four subclasses IgG1-PT, IgG2-PT, IgG3-PT, IgG4-PT)
are removed from antibody inputs at all timepoints (d0, d3, d7, d30) to
avoid proximal label leakage. $\Delta_{t}$: $\log_2$ fold change from
baseline at day $t$.}
\label{tab:design}
{\small
\begin{tabular}{l l l l}
\hline
\textbf{Modality} & \textbf{Input features} & \textbf{Timepoints} & \textbf{d1 used?} \\
\hline
Antibody  & $x_0, \Delta_3, \Delta_7, \Delta_{30}$; PT family removed at all timepoints & d0, d3, d7, d30 & No \\
Cytokine  & $x_0, \Delta_1, \Delta_7, \Delta_{14}$ & d0, d1, d7, d14 & Yes \\
Cell freq & $x_0, \Delta_1, \Delta_3, \Delta_{14}$ & d0, d1, d3, d14 & Yes \\
Gene expr & $x_0, \Delta_7, \Delta_{14}$ & d0, d7, d14 & No \\
\hline
\end{tabular}
}
\end{table}

Two metadata features, childhood vaccine type (wP/aP) and biological sex, are concatenated after fusion (see \emph{Model architecture} below).
Subjects lacking specimens at any of the timepoints required by the four modalities (Table~\ref{tab:design}) were excluded from analysis. After this filter, $n = 158$ for Task~1 and $n = 96$ for Task~2.
\subsection*{Label-feature separation}

Because both tasks are derived from IgG-PT measurements at different
timepoints, care is needed to ensure that no feature used as input also
appears in label construction. We apply three checks during feature
generation.

First, day 14 antibody measurements are excluded from the input
features entirely, since they form the Task~1 numerator
($\log_2$ d14/d0 IgG-PT). Day 120 measurements are never used as
inputs either; they form the Task~2 numerator
($\log_2$ d120/d30 IgG-PT).

Second, pertussis-toxin-specific antibody features (IgG-PT and its four subclasses IgG1-PT, IgG2-PT, IgG3-PT, IgG4-PT) are removed from the antibody input table
\emph{at every input timepoint} (day~0, day~3, day~7, day~30). Although
days 3 and 7 do not directly enter either label, retaining them risks
proximal label leakage because anti-PT IgG titers are autocorrelated
within a subject across days; we therefore exclude the entire PT family
from the antibody inputs. Anti-PT signal is still represented in the model via the d3 and d7
IgG titers of \emph{other} pertussis antigens (FIM2/3, FHA, PRN),
which are also components of the acellular vaccine and which elicit
correlated antibody responses upon boosting~\cite{Edwards2018,
Burdin2017}, but are not direct label constituents.

Third, gene expression preprocessing (variance filtering to the top
2,000 features and standardization) is fit on training data only, then
applied to validation and test data without re-fitting. This prevents
test set statistics from influencing feature selection or scaling.

\subsection*{Model architecture}

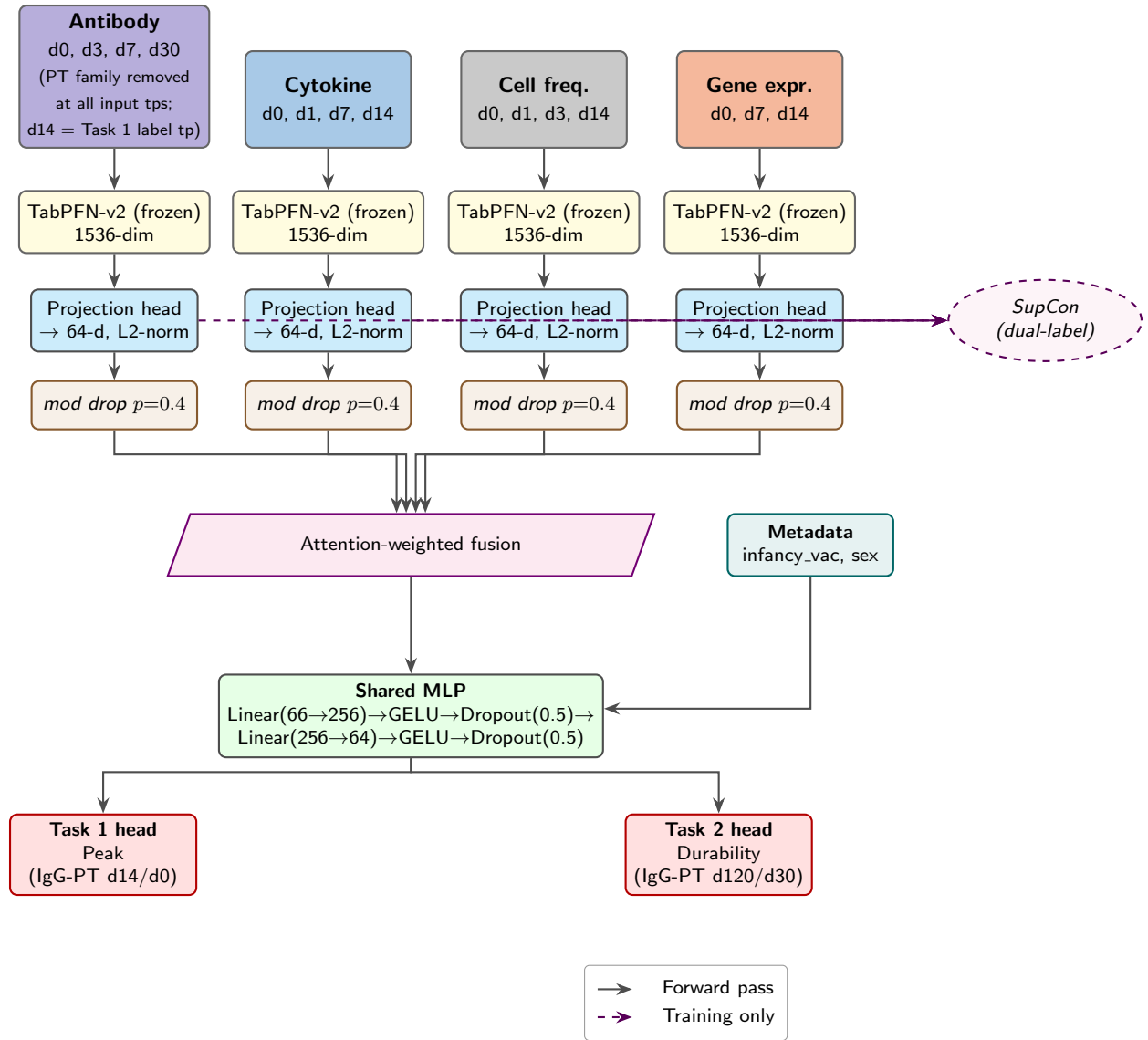
\begin{figure*}[tbp]
\centering
\begin{tikzpicture}[
    font=\sffamily\small,
    >=Stealth,
    node distance=0.5cm,
    modbox/.style={
        rectangle, rounded corners=3pt, draw=black!60, thick,
        minimum width=2.4cm, minimum height=1.4cm, align=center
    },
    tabpfn/.style={
        rectangle, rounded corners=3pt, draw=black!60, thick, fill=yellow!15,
        minimum width=2.4cm, minimum height=0.9cm, align=center,
        font=\sffamily\footnotesize
    },
    proj/.style={
        rectangle, rounded corners=3pt, draw=black!70, thick, fill=cyan!18,
        minimum width=2.4cm, minimum height=0.9cm, align=center,
        font=\sffamily\footnotesize
    },
    dropout/.style={
        rectangle, rounded corners=3pt, draw=brown!70!black, thick, fill=brown!12,
        minimum width=2.4cm, minimum height=0.7cm, align=center,
        font=\sffamily\footnotesize\itshape
    },
    fusion/.style={
        trapezium, trapezium left angle=70, trapezium right angle=110,
        draw=violet!90!black, thick, fill=magenta!10,
        minimum width=3.5cm, minimum height=0.9cm, align=center,
        font=\sffamily\footnotesize
    },
    meta/.style={
        rectangle, rounded corners=3pt, draw=teal!80!black, thick, fill=teal!10,
        minimum width=2.4cm, minimum height=0.9cm, align=center,
        font=\sffamily\footnotesize
    },
    sharedmlp/.style={
        rectangle, rounded corners=3pt, draw=black!70, thick, fill=green!10,
        minimum width=4cm, minimum height=1.2cm, align=center,
        font=\sffamily\footnotesize
    },
    head/.style={
        rectangle, rounded corners=3pt, draw=red!70!black, thick, fill=red!12,
        minimum width=2.7cm, minimum height=1cm, align=center,
        font=\sffamily\footnotesize
    },
    conloss/.style={
        ellipse, draw=violet!70!black, thick, dashed, fill=magenta!5,
        minimum width=2.8cm, minimum height=0.8cm, align=center,
        font=\sffamily\footnotesize\itshape
    },
    arr/.style={->, thick, black!70},
    arrdashed/.style={->, thick, dashed, violet!70!black}
]

\definecolor{geneCol}{HTML}{F5B89C}      
\definecolor{cytoCol}{HTML}{A8C9E7}      
\definecolor{antiCol}{HTML}{B7AEDB}      
\definecolor{cellCol}{HTML}{C9C9C9}      

\node[modbox, fill=cytoCol]                       (cy) {\textbf{Cytokine}\\\footnotesize d0, d1, d7, d14};
\node[modbox, fill=cellCol,  right=0.7cm of cy]   (cl) {\textbf{Cell freq.}\\\footnotesize d0, d1, d3, d14};
\node[modbox, fill=geneCol,  right=0.7cm of cl]   (ge) {\textbf{Gene expr.}\\\footnotesize d0, d7, d14};
\node[modbox, fill=antiCol, anchor=south] (ab) at ([xshift=-3.1cm]cy.south)
   {\textbf{Antibody}\\\footnotesize d0, d3, d7, d30\\
    \scriptsize (PT family removed\\\scriptsize at all input tps;\\\scriptsize d14 = Task 1 label tp)};

\node[tabpfn, below=0.6cm of cy] (tab_cy) {TabPFN-v2 (frozen)\\1536-dim};
\node[tabpfn] at (ab |- tab_cy) (tab_ab) {TabPFN-v2 (frozen)\\1536-dim};
\node[tabpfn] at (cl |- tab_cy) (tab_cl) {TabPFN-v2 (frozen)\\1536-dim};
\node[tabpfn] at (ge |- tab_cy) (tab_ge) {TabPFN-v2 (frozen)\\1536-dim};

\node[proj, below=0.5cm of tab_cy] (p_cy) {Projection head\\$\to$ 64-d, L2-norm};
\node[proj] at (tab_ab |- p_cy) (p_ab) {Projection head\\$\to$ 64-d, L2-norm};
\node[proj] at (tab_cl |- p_cy) (p_cl) {Projection head\\$\to$ 64-d, L2-norm};
\node[proj] at (tab_ge |- p_cy) (p_ge) {Projection head\\$\to$ 64-d, L2-norm};

\node[conloss, right=1.5cm of p_ge] (conloss) {SupCon\\(dual-label)};

\node[dropout, below=0.4cm of p_cy] (d_cy) {mod drop $p{=}0.4$};
\node[dropout] at (p_ab |- d_cy) (d_ab) {mod drop $p{=}0.4$};
\node[dropout] at (p_cl |- d_cy) (d_cl) {mod drop $p{=}0.4$};
\node[dropout] at (p_ge |- d_cy) (d_ge) {mod drop $p{=}0.4$};

\node[fusion, below=1.2cm of d_cy.south, xshift=1.2cm] (fuse)
    {Attention-weighted fusion};
\node[meta, right=1.2cm of fuse] (metadata) {\textbf{Metadata}\\\footnotesize infancy\_vac, sex};

\node[sharedmlp, below=1.4cm of fuse] (shared)
    {\textbf{Shared MLP}\\\footnotesize Linear(66$\to$256)$\to$GELU$\to$Dropout(0.5)$\to$\\\footnotesize Linear(256$\to$64)$\to$GELU$\to$Dropout(0.5)};

\node[head, below left=0.8cm and 0.3cm of shared] (head1)
    {\textbf{Task 1 head}\\\footnotesize Peak\\\footnotesize (IgG-PT d14/d0)};
\node[head, below right=0.8cm and 0.3cm of shared] (head2)
    {\textbf{Task 2 head}\\\footnotesize Durability\\\footnotesize (IgG-PT d120/d30)};

\draw[arr] (ab) -- (tab_ab);  \draw[arr] (cy) -- (tab_cy);
\draw[arr] (cl) -- (tab_cl);  \draw[arr] (ge) -- (tab_ge);
\draw[arr] (tab_ab) -- (p_ab); \draw[arr] (tab_cy) -- (p_cy);
\draw[arr] (tab_cl) -- (p_cl); \draw[arr] (tab_ge) -- (p_ge);

\draw[arrdashed] (p_ab.east) -- ++(0.15,0) |- (conloss.west);
\draw[arrdashed] (p_cy.east) -- ++(0.15,0) |- (conloss.west);
\draw[arrdashed] (p_cl.east) -- ++(0.15,0) |- (conloss.west);
\draw[arrdashed] (p_ge.east) -- (conloss.west);

\draw[arr] (p_ab) -- (d_ab); \draw[arr] (p_cy) -- (d_cy);
\draw[arr] (p_cl) -- (d_cl); \draw[arr] (p_ge) -- (d_ge);

\coordinate (fuse_in_ab) at ($(fuse.north)!-0.45!(fuse.north east)$);
\coordinate (fuse_in_cy) at ($(fuse.north)!-0.15!(fuse.north east)$);
\coordinate (fuse_in_cl) at ($(fuse.north)! 0.15!(fuse.north east)$);
\coordinate (fuse_in_ge) at ($(fuse.north)! 0.45!(fuse.north east)$);

\draw[arr] (d_ab.south) |- ++(0,-0.35) -| (fuse_in_ab);
\draw[arr] (d_cy.south) |- ++(0,-0.35) -| (fuse_in_cy);
\draw[arr] (d_cl.south) |- ++(0,-0.35) -| (fuse_in_cl);
\draw[arr] (d_ge.south) |- ++(0,-0.35) -| (fuse_in_ge);

\draw[arr] (fuse.south) -- (shared.north);
\draw[arr] (metadata.south) |- ([yshift=3pt]shared.east);

\draw[arr] (shared.south) -- ++(0,-0.2) -| (head1.north);
\draw[arr] (shared.south) -- ++(0,-0.2) -| (head2.north);

\node[draw=black!40, thin, rounded corners=2pt, fill=white,
      inner sep=5pt, font=\sffamily\footnotesize,
      below=1.0cm of head2, xshift=1cm, anchor=north east] (legend) {
    \begin{tabular}{@{}ll@{}}
    \tikz[baseline=-0.5ex]\draw[->, thick, black!70] (0,0) -- (0.5,0); & Forward pass \\[2pt]
    \tikz[baseline=-0.5ex]\draw[->, thick, dashed, violet!70!black] (0,0) -- (0.5,0); & Training only \\
    \end{tabular}
};

\end{tikzpicture}
\caption{\textbf{Multi-task contrastive multimodal fusion architecture.}
Four immune modalities are independently encoded by frozen TabPFN-v2
encoders and projected to a shared 64-dim space via per-modality MLP
heads with $l_2$ normalization. A dual-label supervised contrastive loss
(dashed arrows; training only) aligns same-class representations across
modalities. Modality dropout ($p = 0.4$; training only) masks each
modality independently to encourage robustness to missing inputs. An
attention-weighted fusion layer combines present modalities into a
single representation, which is concatenated with subject metadata
(\texttt{infancy\_vac}, biological sex), passed through a shared MLP,
and branched into two classification heads: Task~1 (peak response,
$\log_2$ IgG-PT day~14/day~0) and Task~2 (durability, $\log_2$ IgG-PT
day~120/day~30).}
\label{fig:architecture}
\end{figure*}

The architecture (Fig~\ref{fig:architecture}) is a discriminative
multi-task contrastive multimodal fusion model. It takes per-subject
feature tables from four immune modalities, produces a single fused
64-dimensional representation, and predicts two binary classification
targets jointly. Three design choices address the core challenges at
this sample size: frozen pretrained TabPFN-v2 encoders make each
modality individually informative without requiring per-modality
training on our small dataset; modality dropout and missingness-masked
attention fusion let the model train and infer under arbitrary subsets
of missing modalities; and a dual-label supervised contrastive loss
transfers information across subjects by comparing pairs, despite
having only a few dozen subjects per class.

\subsubsection*{Inputs}
For each modality, the input is a feature table with subject-specific
baseline values and log-fold-change columns at pre-label timepoints:
antibody (day 0, 3, 7, 30; day 14 excluded as the Task~1 label timepoint, and the PT family (IgG-PT and its four subclasses IgG1-PT, IgG2-PT, IgG3-PT, IgG4-PT) removed at all input timepoints to avoid proximal label leakage), cytokine (day
0, 1, 7, 14), cell frequency (day 0, 1, 3, 14), and gene expression
(day 0, 7, 14). Gene expression is pre-filtered by variance to the
top 2,000 features on the training set before encoding. Metadata
(\texttt{infancy\_vac}, biological sex) is held aside as a
two-dimensional auxiliary input.

Modality availability varies across subjects. Antibody is nearly fully
observed (0\% missing); gene expression is also nearly complete (12.7\%
missing); cell frequency and cytokine are more sporadic (27.8\% and
38.6\% missing respectively, in the Task~1 subset). In aggregate,
44.9\% of subjects lack at least one complete modality. Missingness
is at the level of entire modalities, not individual features, and is
driven primarily by cohort-level assay availability rather than random
dropout.

\subsubsection*{Per-modality TabPFN-v2 encoders (frozen)}
Each modality's feature table is passed through a frozen
TabPFN-v2~\cite{TabPFN, TabPFNv2} tabular foundation model, which
returns a 1,536-dimensional in-context embedding per subject. TabPFN-v2
is pre-trained on a large corpus of synthetic tabular tasks and produces
meaningful embeddings for new tables without fine-tuning. We use it as
a frozen feature extractor: the encoder weights do not change during
training. At our sample size (a few dozen subjects per modality after
splitting), training a separate encoder per modality from scratch would
overfit; TabPFN-v2 imports pretrained structure we could not afford to
learn ourselves.

\subsubsection*{Projection heads}
Independent two-layer MLP projection heads $g_m(\cdot)$, one per
modality, map each modality's 1,536-dimensional TabPFN embedding
into a shared $64$-dimensional space (hidden dimension $256$),
followed by $\ell_2$-normalization:
\begin{equation}
\boldsymbol{h}_m^{(i)} = \frac{g_m(\boldsymbol{z}_m^{(i)})}{\|g_m(\boldsymbol{z}_m^{(i)})\|_2},
\qquad g_m(\boldsymbol{z}) = W_2^{(m)} \, \sigma\!\left(\mathrm{LN}(W_1^{(m)} \boldsymbol{z} + \boldsymbol{b}_1^{(m)})\right) + \boldsymbol{b}_2^{(m)}
\end{equation}
where $\sigma$ is GELU activation and LN is layer normalization
(dropout between activation and the second linear layer is omitted
for clarity; see Table~\ref{tab:hyperparams}). There is no weight
sharing across modalities: antibody, cytokine, cell frequency, and
gene expression carry fundamentally different biological information,
and a per-modality projection preserves modality-specific structure
that a shared head would average over. The $\ell_2$ normalization
places the projected embeddings on the unit sphere, which is the
geometric setting required by the supervised contrastive loss
below~\cite{SupCon}.

\subsubsection*{Dual-label supervised contrastive loss}
A supervised contrastive (SupCon) loss~\cite{SupCon} operates on the
normalized projections $\boldsymbol{h}_m^{(i)}$ during training only
(dashed arrows in Fig~\ref{fig:architecture}). The loss pulls together
the embeddings of subjects designated as a \emph{positive pair} and
pushes apart those of \emph{negative pairs}. We extend single-label
SupCon to a \emph{dual-label} variant: two subjects form a positive pair
if they agree on the Task~1 label \emph{or} the Task~2 label, so both
tasks have equal influence on the learned geometry. When the Task~2
label is missing (2023 cohort, encoded as $-1$), subjects participate
in positive-pair construction via their Task~1 label alone. Temperature
$\tau = 0.3$; contrastive weight $\lambda = 0.1$. We use the ``OR''
definition instead of ``AND'' because Task~1 and Task~2 labels are
strongly anti-correlated (Spearman $r = -0.58$), so requiring agreement
on both would produce too few positive pairs per batch to train a
stable contrastive objective.

\subsubsection*{Modality dropout}
During training, each of the four modalities is independently zeroed
with probability $p = 0.4$ on each forward pass, with the constraint
that at least one modality is always retained. Dropout is applied per
subject per step, so the same subject is seen through different modality
subsets across different training steps. Over an epoch, the model is
exposed to a wide range of modality combinations. The per-modality
rate $p = 0.4$ is set in the same range as the empirical per-modality
missingness rates in the Task~1 subset, which lie between $0\%$
(antibody) and $38.6\%$ (cytokine). Note that under independent per-modality dropout at $p = 0.4$, the
probability that at least one modality is dropped on a given forward
pass is $1 - (1-p)^4 = 1 - 0.6^4 = 0.87$ (slightly lower in practice because the all-dropped case ($p^4 = 0.026$) is prevented). This exceeds the $44.9\%$ rate
at which subjects in the Task~1 subset are missing at least one
modality. The training time dropout regime is therefore deliberately more aggressive than the inference time missingness pattern, in order to encourage robustness to modality combinations that may not be represented in the training set.
The purpose is to ensure that any modality combination encountered at inference, including subjects missing any subset of the four modalities, is in-distribution for the trained model rather than a distribution shift.
Modality dropout is switched off at inference; practical missingness at test
time is handled by the attention fusion below.

\subsubsection*{Attention-weighted fusion with missingness masking}
A learned query vector $\boldsymbol{w} \in \mathbb{R}^{64}$ computes
attention scores over the present modalities for each subject.
Critically, modalities that are absent (whether due to real missingness
or modality dropout during training) are excluded from the softmax
computation, so that the attention weights are computed and normalized
only over the subset of modalities that are actually observed:
\begin{equation}
\alpha_m^{(i)} =
\frac{\exp(\boldsymbol{w}^\top \boldsymbol{h}_m^{(i)})}
     {\sum_{m': \mu_{m'}^{(i)}=1} \exp(\boldsymbol{w}^\top \boldsymbol{h}_{m'}^{(i)})}
\quad \text{for } m \text{ with } \mu_m^{(i)} = 1,
\qquad
\boldsymbol{f}^{(i)} = \sum_{m=1}^{M} \mu_m^{(i)} \cdot \alpha_m^{(i)} \cdot \boldsymbol{h}_m^{(i)}
\end{equation}
where $\mu_m^{(i)} \in \{0, 1\}$ indicates whether modality $m$ is
present for subject $i$. The weights $\alpha_m^{(i)}$ sum to one over
the observed modalities, so the fused representation is a convex
combination of the available modalities with no imputation or
placeholder values. The same fusion code handles training (with
dropout-induced missingness) and inference (with real missingness)
without any mode switch.

\subsubsection*{Metadata injection into the shared MLP}
The 64-dimensional fused representation $\boldsymbol{f}^{(i)}$ is
concatenated with a 2-dimensional binary metadata vector
(childhood priming and biological sex), yielding a 66-dimensional
input to the shared MLP:
Linear$(66 \to 256) \to \mathrm{GELU} \to \mathrm{Dropout}(0.5) \to
\mathrm{Linear}(256 \to 64)$. Metadata is injected after fusion rather
than as a fifth fusion modality. This choice has an important
interpretability consequence: because the attention layer never sees
\texttt{infancy\_vac}, it cannot learn to route modality attention based
on priming status.

\subsubsection*{Task heads}
Two linear classification heads branch from the shared MLP output, each
producing 2-class logits: Task~1 (peak response) and Task~2
(durability). Each head is a single Linear$(64 \to 2)$ with no
head-specific hidden layers. Keeping the heads linear forces
task-discriminative structure to live in the shared MLP, where it is
constrained to also serve the other task. A non-linear per-task head
would let each task silently build a specialized representation that
diverges from the shared one, which would defeat the purpose of
multi-task learning.

\subsubsection*{Joint training objective}
Total loss is
\begin{equation}
\mathcal{L} = \mathcal{L}_{\mathrm{CE}}^{(\mathrm{T1})}
            + w_{\mathrm{T2}} \cdot \mathcal{L}_{\mathrm{CE}}^{(\mathrm{T2})}
              \cdot \mathbf{1}[y_{\mathrm{T2}} \neq -1]
            + \lambda \cdot \mathcal{L}_{\mathrm{con}}
\end{equation}
with $w_{\mathrm{T2}} = 2.0$ and $\lambda = 0.1$. The Task~2
cross-entropy is masked for subjects without day-120 measurements (2023
cohort) but those subjects still contribute to Task~1 cross-entropy
and to the dual-label SupCon loss via their Task~1 label. Optimization
uses AdamW~\cite{AdamW}: learning rate $10^{-2}$, weight decay $10^{-3}$,
cosine annealing schedule, gradient clipping at norm $1.0$, batch size
$32$, up to $60$ epochs, early stopping on mean validation AUROC with
patience $10$. The Task~2 loss is up-weighted ($w_{\mathrm{T2}} = 2$)
to compensate for the missing-label masking that excludes the 2023
cohort from the Task~2 cross-entropy term, which would otherwise
underrepresent the durability gradient signal during training. The choice is validated empirically against $w_{\mathrm{T2}} = 1$ in Table~\ref{tab:ablation}.

Full hyperparameter configuration in Table~\ref{tab:hyperparams}.

\subsubsection*{Data split}
We use a fixed stratified train/validation/test split (fixed seed)
yielding $n_{\text{train}} = 94$, $n_{\text{val}} = 32$, and
$n_{\text{test}} = 32$ subjects for Task~1. For Task~2 (durability),
the test set contains $n = 21$ subjects with non-missing retention
labels. We use the hyperparameters in Table~\ref{tab:hyperparams} throughout.
Earlier exploratory Optuna searches across multiple seeds and
hyperparameter ranges did not produce a configuration that consistently
outperformed hand-chosen defaults on validation, likely because the
validation set ($n = 32$) is too small to reliably distinguish small
performance differences between configurations. We therefore did not
re-run a full hyperparameter search on the final feature design and
report results at the default configuration shown.

\begin{table}[!ht]
\centering
\caption{{\bf Hyperparameter configuration.}}
\label{tab:hyperparams}
\begin{tabular}{l l}
\thickhline
\textbf{Hyperparameter} & \textbf{Value} \\
\hline
Projection dim / hidden & 64 / 256 \\
Shared MLP hidden dims & (256, 64) \\
Dropout & 0.5 \\
Contrastive weight ($\lambda$) / temperature ($\tau$) & 0.1 / 0.3 \\
Modality dropout ($p$) & 0.4 \\
Task~2 weight ($w_{\text{T2}}$) & 2.0 \\
Learning rate / weight decay & $10^{-2}$ / $10^{-3}$ \\
Batch size / max epochs / patience & 32 / 60 / 10 \\
\thickhline
\end{tabular}
\end{table}

\subsection*{Experimental design}

\subsubsection*{Permutation test}
To assess whether predictive performance is distinguishable from chance,
we perform a label-shuffling permutation test~\cite{Ojala2010} applied
jointly to both tasks. For each of $N = 1000$ permutations, we
(i)~randomly shuffle the Task~1 labels across all subjects in the full
cohort, (ii)~independently shuffle the Task~2 labels across subjects
that have them (preserving the Task~2 missingness pattern), (iii)~retrain
the full pipeline from scratch on the shuffled dataset using the same fixed seed train/validation/test split and the same hyperparameters, and
(iv)~record the resulting test-set AUROC for both tasks. These $N$
AUROCs form the null distribution: the performance the architecture
would achieve if the feature--label relationship were random. The
one-sided $p$-value for each task is
\begin{equation}
p = \frac{1 + \sum_{i=1}^{N} \mathbf{1}[\text{null}_i \geq \text{obs}]}
        {N + 1}
\end{equation}
where $\text{obs}$ is the observed test AUROC on the real, unshuffled
labels.

Retraining from scratch on each permuted dataset ensures that the
observed AUROC reflects a genuine, learnable feature--label relationship
rather than a quirk of the split, hyperparameter choice, or
initialization. Joint shuffling of both task labels preserves the
validity of the null under the dual-label supervised contrastive loss,
which otherwise would leak information from the unshuffled task during
training.

\subsubsection*{Bootstrap confidence intervals}
To quantify uncertainty on the observed test AUROCs, we compute
bootstrap confidence intervals by resampling the test set with
replacement. For each of $B = 1000$ resamples, we sample
$n_{\text{test}}$ subject indices uniformly with replacement from the
held-out test set, compute AUROC on the resampled subjects using the
trained model's fixed predictions (without retraining), and record the
value. Resamples in which the bootstrap sample contains only a single
class (degenerate AUROC) are discarded. The $95\%$ confidence interval
is then the $2.5$th and $97.5$th percentiles of the resulting bootstrap
distribution (percentile method). The architecture is evaluated on a single seed train/val/test split. Although the bootstrap confidence intervals quantify evaluation uncertainty given this split, they do not capture variability due to split selection or training stochasticity; multi-seed evaluation is deferred to subsequent work.

\subsubsection*{Per-modality contribution analyses}
We use two complementary analyses to quantify what each modality
contributes. \emph{Leave-one-out (LOO):} we recompute test AUROC with
one modality masked out at inference (its mask set to zero, attention
renormalized over the remaining three), repeated for each modality.
\emph{Keep-one-out (KOO):} we recompute test AUROC keeping only one
modality at inference, with the other three masked out. Both are
conducted on the trained preferred model without retraining; the model
itself is unchanged. LOO measures \emph{additive contribution} (how much
worse without modality $m$); KOO measures \emph{standalone sufficiency}
(how well modality $m$ alone supports prediction). Together they
diagnose whether a modality is necessary, sufficient, both, or neither.

\subsubsection*{Graceful degradation}
To probe robustness to modality loss at deployment, we simulate a
range of missingness severities at inference: for each modality,
we randomly mask its values for a fraction
$\rho \in \{0.0, 0.1, 0.2, 0.3, 0.5, 0.7, 1.0\}$ of test subjects
and recompute test AUROC. The masking pattern is drawn under a
fixed seed. A graceful degradation profile shows how AUROC
varies with $\rho$, indicating how the missingness-aware fusion
handles practical deployment where any subset of subjects might
lack any subset of modalities.

\subsubsection*{Architectural ablation}
We evaluate five configurations on the same train/val/test split as
the main results: (i)~\emph{Full (preferred)} with contrastive loss
($\lambda = 0.1$), modality dropout ($p = 0.4$), and Task~2
upweighting ($w_{\mathrm{T2}} = 2$); (ii)~\emph{No contrastive}
($\lambda = 0$) which removes the dual-label SupCon alignment;
(iii)~\emph{No modality dropout} ($p = 0$) which disables the
training-time random masking of modalities, so each subject is
seen during training only with the modality combination they
were measured under;
(iv)~\emph{Neither}, with both contrastive and modality dropout
removed; and (v)~\emph{No T2 up-weighting} ($w_{\mathrm{T2}} = 1$)
which removes the durability-loss compensation. The same seed is used throughout and the same hyperparameters otherwise. We report test AUROC and
bootstrap 95\% CI for both tasks.

\subsection*{Baseline comparisons}

For benchmarking the multi-task fusion architecture,
we trained three baseline classifiers on the concatenation of raw
features from all four modalities (antibody, cytokine, cell frequency,
gene expression): logistic regression from
scikit-learn~\cite{scikit-learn}, gradient-boosted trees from
XGBoost~\cite{xgboost}, and a small multilayer perceptron
(\emph{TabMLP}: two hidden layers of size $128$ and $64$ with GELU
activation and dropout $0.3$, trained with AdamW, cross-entropy loss,
batch size $32$, learning rate $10^{-3}$, weight decay $10^{-4}$, for
$150$ epochs). Missing modality values were imputed using per-feature
training-set means; logistic regression and TabMLP additionally used
training-set z-score standardisation, while XGBoost was trained on
imputed-but-unstandardised features per its standard usage. All three
baselines used the same seed as the rest of the code with the same train/test split as well, and the default hyperparameters (XGBoost: 100 trees, max
depth 3, learning rate 0.1, histogram tree method; logistic
regression: $L_2$ penalty, $C = 1.0$, LBFGS solver, max 2000
iterations). Task~2 baselines were fit only on subjects with a
non-missing durability label. Test-set AUROC confidence intervals were
computed by the same 1000-resample bootstrap used for the main
results.

To isolate the contribution of the multi-task contrastive fusion
architecture from the contribution of the underlying TabPFN-v2
feature extractor, we additionally evaluated the three baseline
classifiers on concatenated TabPFN-v2 embeddings. For each subject,
we generated a 1,536-dimensional TabPFN embedding per modality
(using the same per-modality feature design as the main model) and
concatenated them into a single 6,144-dimensional input vector,
with subjects missing a modality assigned a zero vector for that
modality's embedding block. Logistic regression, XGBoost, and
TabMLP were then trained on these concatenated embeddings using
the same hyperparameters and train/test split as the raw-feature
baselines.

\section*{Results}
\subsection*{Peak and durability are anti-correlated}

\begin{figure*}[tbp]
\centering
\includegraphics[width=0.72\textwidth]{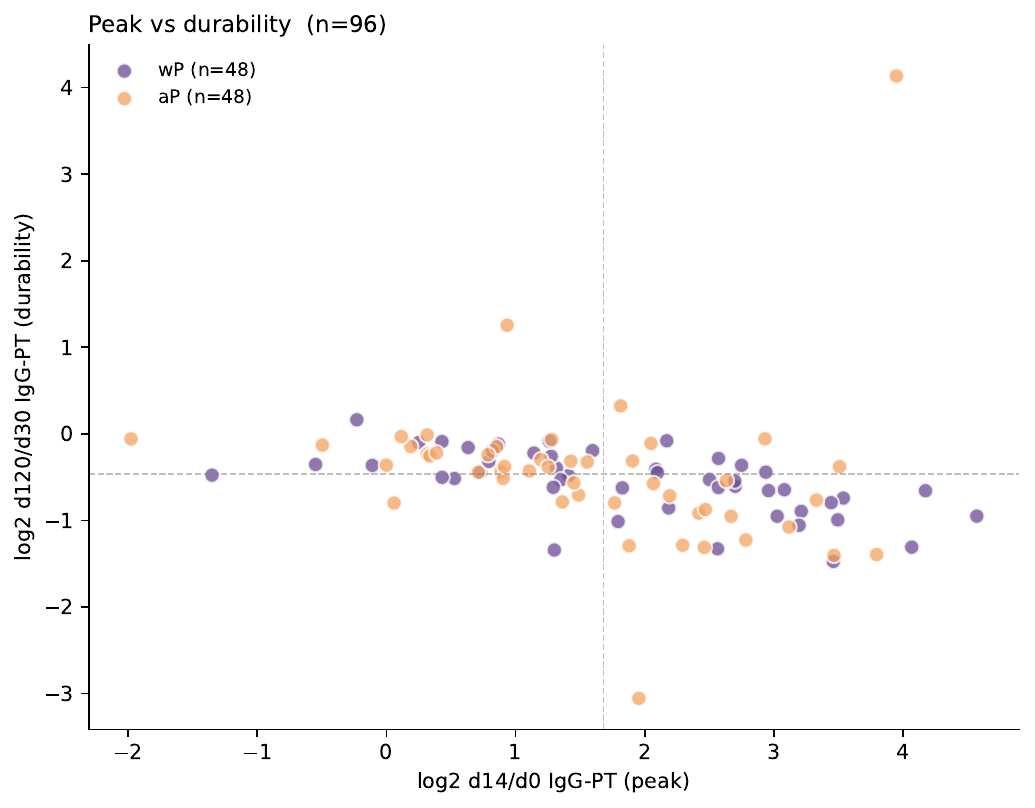}
\caption{\textbf{Peak and durability are anti-correlated phases of the
humoral response.} Scatter of $\log_2$ day~120/day~30 retention (durability, y-axis)
versus $\log_2$ day~14/day~0 IgG-PT fold change (peak, x-axis) for the subjects with both labels available ($n = 96$).
Subjects colored by childhood vaccination history. Dotted vertical line:
peak median used to derive the Task~1 binary label. Dashed horizontal
line: durability median used to derive the Task~2 binary label. The strong negative correlation (Spearman $r = -0.58$,
$p = 4.0 \times 10^{-10}$; Cohen's $\kappa = -0.52$) reflects
the dissociation between acute and long-term phases of the
humoral response, motivating the multi-task framing. Childhood priming is significantly associated with the Task~1 binary
label ($2 \times 2$ $\chi^2$ on priming $\times$ peak: $p = 0.017$,
OR $\approx 2.27$) but does not by itself separate responders from
non-responders, motivating the use of multimodal immune measurements
on top of priming history as model inputs.}
\label{fig:motivation}
\end{figure*}

Peak fold change and durability retention are strongly negatively
correlated (Spearman $r = -0.580$, $p = 3.97 \times 10^{-10}$,
$n = 96$; Cohen's $\kappa = -0.520$ on the binary labels;
Fig~\ref{fig:motivation}). The $2 \times 2$ quadrant counts confirm
the asymmetry: $44$ subjects fall in high peak / low durability,
$30$ in low peak / high durability, $20$ in high peak / high
durability, and only $4$ in low peak / low durability. The two
tasks are statistically dissociated rather than redundant.

\subsection*{Predictive performance on both tasks}

\begin{figure*}[tbp]
\centering
\includegraphics[width=0.95\textwidth]{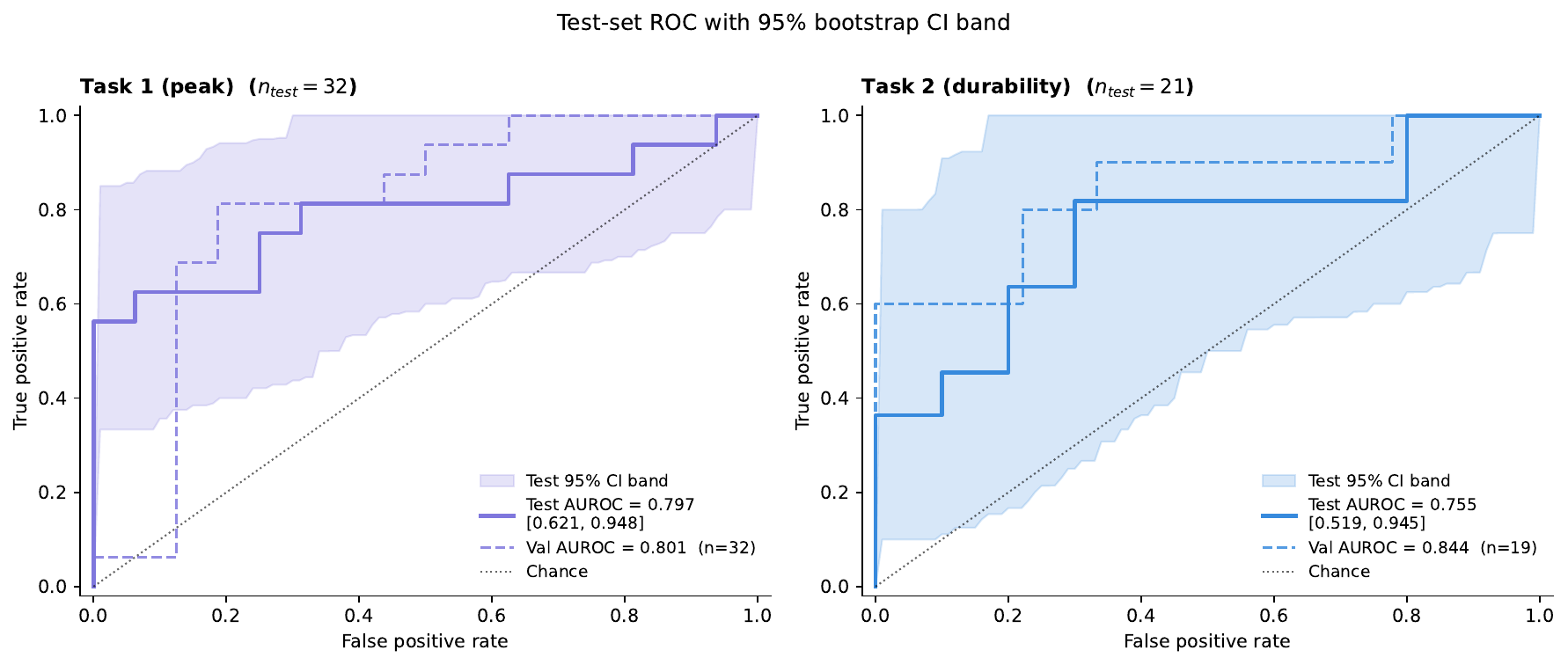}
\caption{\textbf{ROC curves on the held-out test set with 95\% bootstrap
confidence intervals, and validation-set ROCs for comparison.}
\textbf{Left:}~Task~1 (peak response, $n_{\text{test}} = 32$); test
AUROC $0.797$ (95\% CI $[0.621, 0.948]$), validation AUROC $0.801$.
\textbf{Right:}~Task~2 (durability, $n_{\text{test}} = 21$); test AUROC
$0.755$ (95\% CI $[0.519, 0.945]$), validation AUROC $0.844$. Solid
line: ROC on the observed test set. Shaded band: pointwise 95\%
confidence interval on the true positive rate at each false positive
rate, from $B = 1000$ bootstrap resamples of the test set. Dashed
line: validation-set ROC on the same trained model. Dotted diagonal:
chance. Validation and test AUROC agree closely on Task~1 ($0.801$
vs $0.797$). On Task~2 the validation AUROC ($0.844$) exceeds the test AUROC
($0.755$) by approximately $0.09$. Given the small Task~2 test
set ($n = 21$) and the use of validation AUROC for early stopping,
this gap is consistent with mild optimism in validation-based model
selection rather than a clear failure of generalization.}
\label{fig:roc}
\end{figure*}

\begin{figure*}[tbp]
\centering
\includegraphics[width=0.95\textwidth]{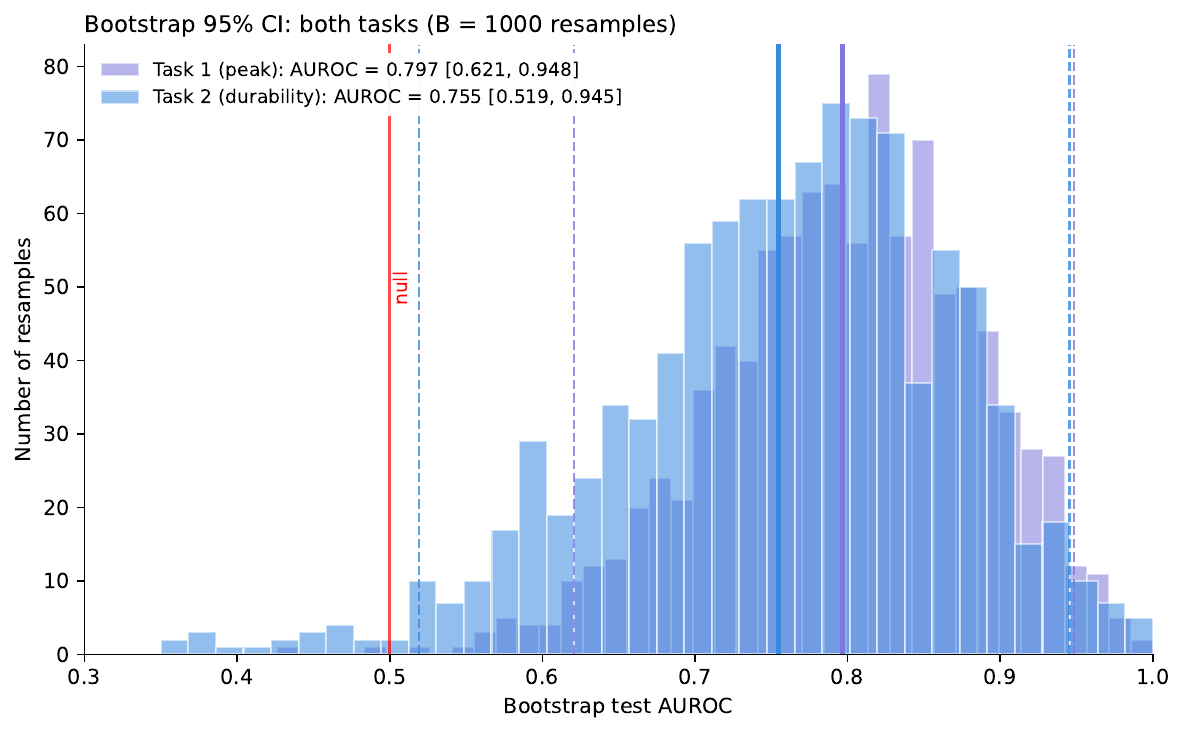}
\caption{\textbf{Bootstrap AUROC distributions on the held-out test
set, both tasks overlaid.} Histograms of test-set AUROC across $B = 1000$
bootstrap resamples (sampling subjects with replacement from the
held-out test set, predictions held fixed). Task~1 (peak response,
lavender): observed AUROC $0.797$, 95\% CI $[0.621, 0.948]$. Task~2
(durability, blue): observed AUROC $0.755$, 95\% CI $[0.519, 0.945]$.
Solid coloured vertical lines: observed AUROC for each task. Dashed
coloured vertical lines: $2.5$th and $97.5$th percentiles of the
bootstrap distribution, defining the 95\% percentile-method confidence
interval. Red vertical line: null AUROC ($0.5$). The Task~2 distribution
is wider than Task~1, reflecting the smaller test set ($n = 21$
vs $n = 32$).}
\label{fig:bootstrap}
\end{figure*}

The preferred configuration achieves test AUROC $0.797$ for Task~1 and
$0.755$ for Task~2 (Fig~\ref{fig:roc}). Bootstrap 95\% confidence
intervals from 1000 resamples are $[0.621, 0.948]$ for Task~1 and
$[0.519, 0.945]$ for Task~2; the full bootstrap distributions are
shown in Fig~\ref{fig:bootstrap}. The Task~2 CI is comparably wide
despite the smaller test set ($n = 21$ versus $n = 32$ for Task~1)
because the durability signal is inherently noisier.

\begin{figure*}[tbp]
\centering
\includegraphics[width=0.95\textwidth]{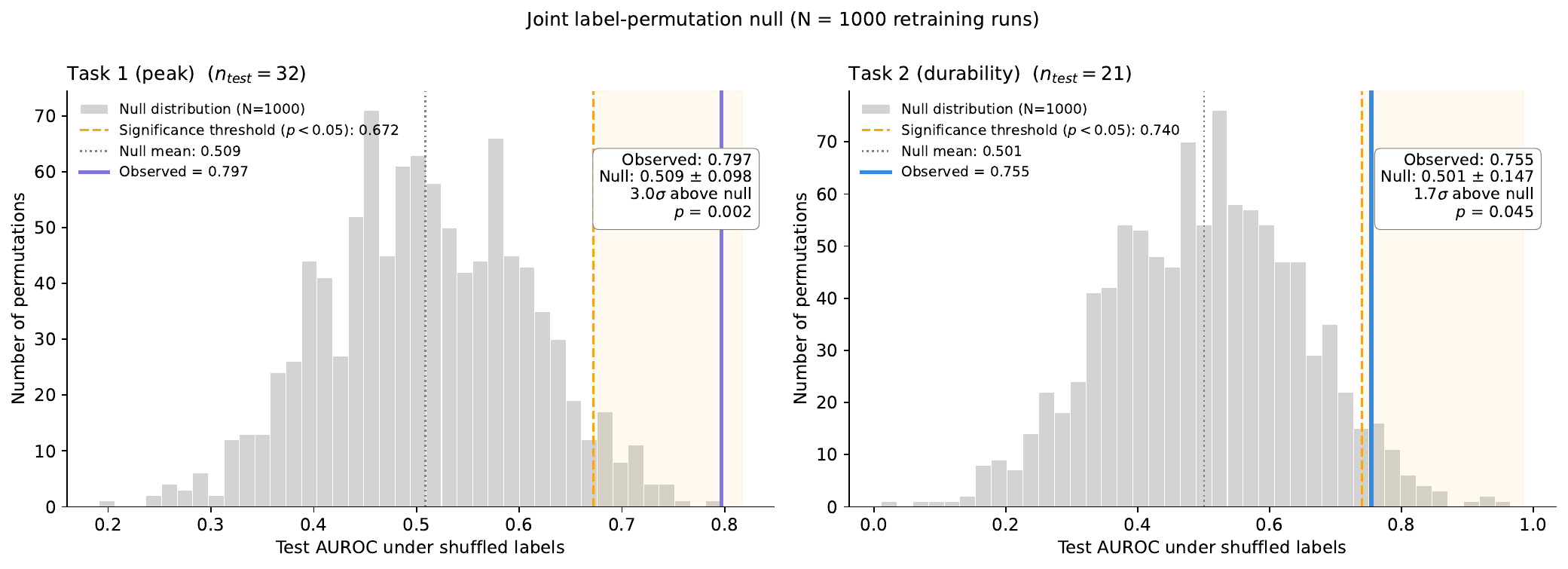}
\caption{\textbf{Label-permutation null distributions confirm learned
signal for both tasks.} Histograms of test-set AUROC across $N = 1000$
joint label permutations (both Task~1 and Task~2 labels shuffled; full
pipeline retrained per permutation). The dashed orange line marks the
significance threshold (95th percentile of the null, $p < 0.05$); the
rejection region is shaded lightly. The dotted grey line marks the
null mean. \textbf{Left:}~Task~1 (peak response); observed AUROC
$0.797$ exceeds the null at $p = 0.002$, approximately $3.0$ null-SDs
above the null mean. \textbf{Right:}~Task~2 (durability); observed
AUROC $0.755$ exceeds the null at $p = 0.045$, approximately $1.7$
null-SDs above the null mean. Null means (Task~1: $0.509$; Task~2:
$0.501$) sit near chance, confirming the permutation procedure is
well-calibrated.}
\label{fig:permutation}
\end{figure*}

To verify these AUROCs reflect a genuine feature--label relationship
rather than artifacts of the held-out split, we performed a joint label
permutation test with $N = 1000$ retraining runs (Methods). The null
distributions centered near chance (Task~1: mean $0.509$, SD $0.098$;
Task~2: mean $0.501$, SD $0.147$), confirming the architecture does
not produce above-chance performance in the absence of a learnable
signal. The observed Task~1 AUROC of $0.797$ exceeded all but a small
fraction of null permutations ($p = 0.002$), sitting approximately
$3.0$ null-SDs above the null mean (Fig~\ref{fig:permutation}). The
observed Task~2 AUROC of $0.755$ exceeded the null at $p = 0.045$,
approximately $1.7$ null-SDs above the null mean. The wider Task 2 null distribution reflects the smaller labeled test subset; combined with the greater intrinsic difficulty of long-term retention prediction, this yields a less stringent significance result for Task~2 ($p = 0.045$) than for Task~1 ($p = 0.002$). Both results
are significant at $\alpha = 0.05$, with Task~1 additionally
significant at $\alpha = 0.01$.

\subsection*{Per-modality contribution: peak and durability rely on different signals}
\begin{figure*}[tbp]
\centering
\includegraphics[width=0.95\textwidth]{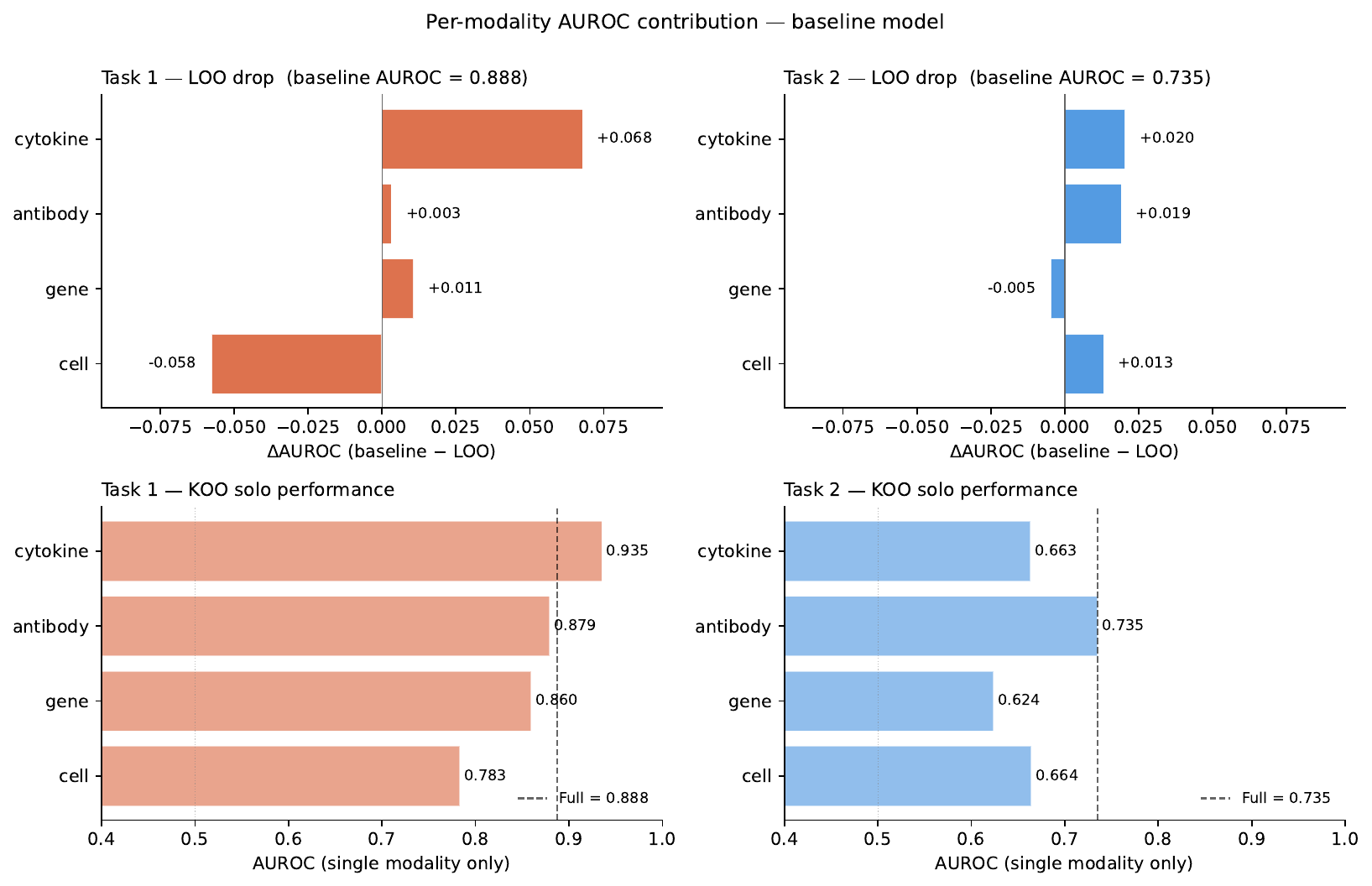}
\caption{\textbf{Per-modality contribution to peak and durability
prediction.} \textbf{Top:}~Leave-one-out (LOO) analysis: change in
test AUROC when each modality is masked out at inference, computed
as $\Delta = \text{baseline AUROC} - \text{LOO AUROC}$. Positive
values indicate the modality contributes to baseline performance;
negative values indicate the modality is harmful to the ensemble.
\textbf{Bottom:}~Keep-one-out (KOO) analysis: test AUROC when only
one modality is retained at inference. High values indicate
standalone sufficiency. All-modalities AUROC is shown as a dashed
reference line. For Task~1, cytokine dominates on both LOO ($+0.068$)
and KOO ($0.935$), while cell frequency hurts the ensemble
(LOO $-0.058$). For Task~2, antibody alone matches the full ensemble
(KOO $= 0.735$). Both analyses are evaluated on the complete-case
test subset.}
\label{fig:modality_contribution}
\end{figure*}

To diagnose what each modality contributes to the results, we conducted complementary
leave-one-out (LOO) and keep-one-out (KOO) analyses on the trained
model (Methods, Fig~\ref{fig:modality_contribution}). Both analyses
are evaluated on the complete-case subset of the test set
(subjects with all four modalities measured), giving a reference
test AUROC of $0.888$ for Task~1 and $0.735$ for Task~2; deviations
from these references quantify modality contribution.

For Task~1 (peak response), the LOO and KOO analyses agree:
\emph{cytokine} is the dominant modality. It is the strongest
single-modality predictor (KOO $= 0.935$, exceeding the
all-modalities value of $0.888$) and the modality whose removal
most harms the ensemble (LOO $\Delta = +0.068$). \emph{Cell
frequency} shows the opposite pattern: it is the weakest
single-modality predictor (KOO $= 0.783$) and removing it from
the ensemble \emph{improves} test AUROC by $0.058$ (LOO
$\Delta = -0.058$), indicating that cell-frequency features add
noise rather than signal at this sample size. Antibody and gene
contribute modestly on both metrics (KOO $0.879$ and $0.860$;
LOO $+0.003$ and $+0.011$).

For Task~2 (durability), the picture is qualitatively different.
\emph{Antibody} alone matches the full-ensemble performance
(KOO $= 0.735$); cytokine, cell, and gene drop sharply in
isolation (KOO $0.663$, $0.664$, $0.624$). LOO drops are uniformly
small (all within $\pm 0.02$ of baseline), indicating that no
single modality is strictly necessary, but antibody clearly
carries the bulk of the durability signal.

This task-specific pattern emerges without any explicit
supervision telling the model which modalities to weight for which
task. The shared MLP and the joint contrastive loss together produce a
representation in which different modality combinations are
diagnostic for different downstream questions, consistent with
the biology of peak response (driven by early innate cytokine
signatures, which predict subsequent antibody outcomes across
vaccines~\cite{Pulendran2014, Querec2009}) versus durability (an
antibody-centric long-term outcome sustained by long-lived plasma
cells~\cite{Amanna2007}).

\subsection*{Graceful degradation under inference-time missingness}

\begin{figure*}[tbp]
\centering
\includegraphics[width=0.95\textwidth]{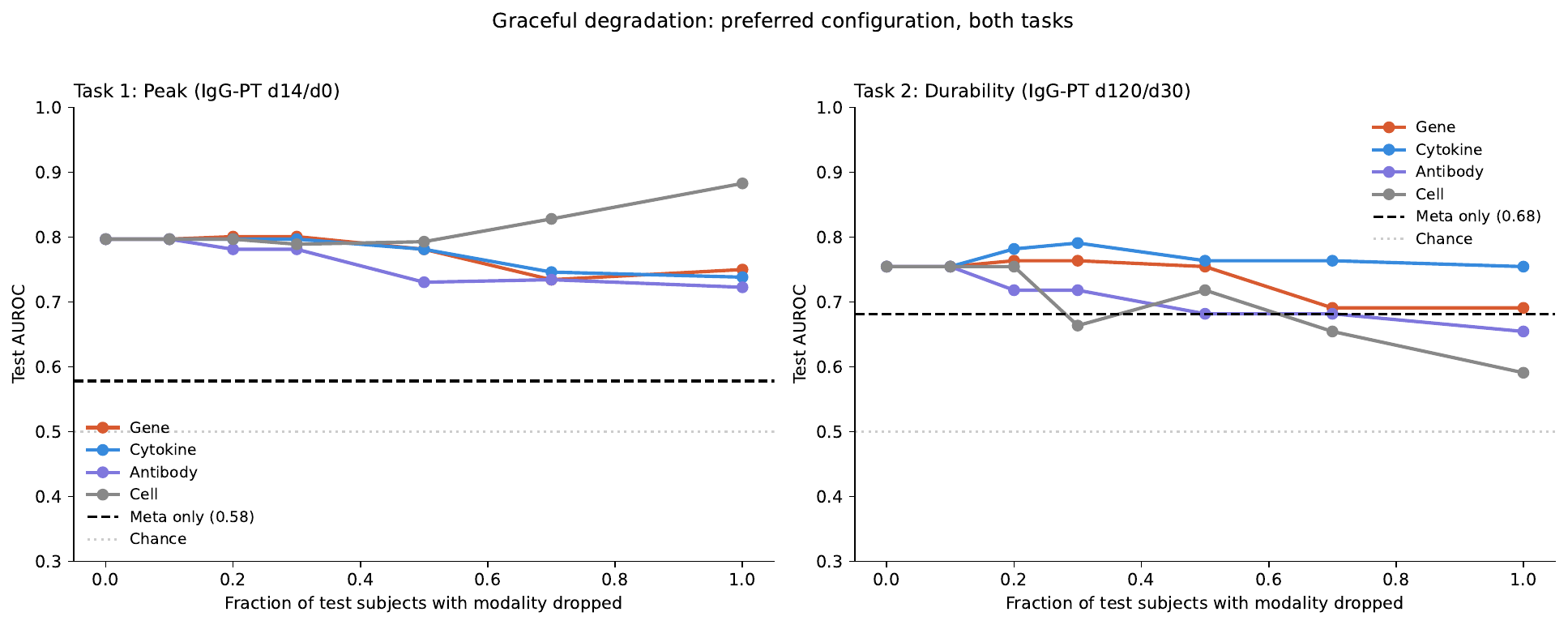}
\caption{\textbf{Graceful degradation under inference-time modality
missingness.} For each modality, a fraction $\rho$ of test subjects
have that modality randomly masked at inference, with attention
renormalised over the remaining three. Test AUROC is computed at each
$\rho \in \{0.0, 0.1, 0.2, 0.3, 0.5, 0.7, 1.0\}$ under a fixed masking
seed (13). \textbf{Left:}~Task~1 (peak response). \textbf{Right:}~Task~2
(durability). Dashed black line: meta-only baseline (the model's AUROC
when all four modalities are masked, leaving only \texttt{infancy\_vac}
and biological sex). Dotted grey line: chance ($0.5$). On Task~1, all
modality curves stay well above meta-only across the full range of
$\rho$. On Task~2, only cytokine remains clearly above meta-only at
$\rho = 1$; gene meets the meta-only line, while antibody and cell
fall below it, with cell crossing meta-only by $\rho \approx 0.3$.}
\label{fig:degradation}
\end{figure*}

Real-world deployment of multimodal models faces an unavoidable
problem: new subjects may arrive with arbitrary subsets of modalities
missing due to assay availability, sample quality, or cost. To probe
how the trained model handles this situation, we simulated a range of
inference-time modality-missingness rates $\rho$ for each of the four
modalities (Methods, Fig~\ref{fig:degradation}).

The trained model handles inference-time modality missingness with
task-asymmetric tolerance. On Task~1, AUROC remains within
$\sim 0.02$--$0.03$ of the all-modalities-present value across the
full range of $\rho$ for every modality, and all curves stay well
above the meta-only baseline ($0.58$) even at $\rho = 1$;
cell-frequency masking on Task~1 actually shows a slight upward
trend rather than a decline (from $0.797$ at $\rho = 0$ to $0.88$
at $\rho = 1$). On Task~2 the safe-tolerance regime is narrower:
AUROC stays close to (or slightly above) the all-modalities-present
value as long as no more than $\sim 20\%$ of test subjects are missing
any one modality. Beyond that, modalities separate sharply. Cytokine
missingness is the most benign: it is in fact slightly beneficial
in the $\rho \approx 0.2$--$0.3$ range, where the curve rises
modestly above the all-modalities-present value before settling
back to $\sim 0.755$ at $\rho = 1$, consistent with cytokine
carrying limited durability-specific signal in the ensemble
(Fig~\ref{fig:modality_contribution}). Gene reaches the meta-only
baseline at $\rho = 1$ ($\sim 0.69$). Antibody and cell drop below
the meta-only baseline ($0.66$ and $0.59$ respectively at
$\rho = 1$); cell-frequency masking is the most damaging, ending
clearly below meta-only at $\rho = 1$. In practical terms, a
deployed model can be safely used on cohorts where any single
modality is missing for up to $\sim 30\%$ of subjects on Task~1
or $\sim 20\%$ on Task~2; cell-frequency missingness in particular
is the limiting deployment factor on durability prediction.

\subsection*{Architectural ablation}
To quantify which architectural components matter, we trained four
configurations of the model on the same fixed seed split: \emph{Full
(preferred)} with both contrastive loss and modality dropout active;
\emph{No contrastive} ($\lambda = 0$); \emph{No modality dropout}
($p = 0$); and \emph{Neither} (both removed). Test AUROC and 95\%
bootstrap CIs are shown in Table~\ref{tab:ablation}.

Both architectural components contribute substantively. Removing the
contrastive loss drops Task~1 AUROC by $0.231$ (from $0.797$ to
$0.566$), the largest single-component effect. Removing modality
dropout drops Task~1 AUROC by $0.176$ (from $0.797$ to $0.621$).
Notably, the configuration with neither component does not produce the worst result on Task~1: at $0.719$, ``Neither'' is in fact \emph{better} than either single-ablation configuration (``No contrastive'' $0.566$, ``No modality dropout'' $0.621$). This
non-monotone pattern indicates that contrastive loss and modality
dropout interact: combining contrastive alignment with full-modality
training (the ``No modality dropout'' case) yields a poorly conditioned
representation that the contrastive loss further sharpens in an
unhelpful direction; removing both components reverts to a less
constrained but more robust configuration. The \emph{Full} configuration,
with both components active, is the only one that achieves AUROC above
$0.79$ on Task~1.

Task~2 results are flatter across configurations (range $0.609$ to
$0.755$), with the \emph{Full} configuration again the best. The
relative robustness of Task~2 to ablation is consistent with its
heavier reliance on the antibody modality alone (see
Fig~\ref{fig:modality_contribution}): when one component is removed,
the antibody pathway through the shared MLP can still carry most of the
durability signal.

A separate row in Table~\ref{tab:ablation} probes the Task~2 loss
weighting. At $w_{\mathrm{T2}} = 1$ (no up-weighting), Task~1 AUROC
drops to $0.668$ and Task~2 to $0.673$, declines of $0.129$ and
$0.082$ respectively. The Task~1 effect is the larger one, which is
the signature of multi-task transfer: properly weighting Task~2
during training yields a better-conditioned shared representation
that benefits Task~1 prediction. This is direct evidence that
multi-task co-training contributes to peak prediction at this
sample size, complementing the contrastive-loss and modality-dropout
ablations above.
\begin{table}[!ht]
\centering
\caption{{\bf Architectural ablation: test-set AUROC with 95\%
bootstrap CIs.} The full configuration includes both the dual-label
contrastive loss ($\lambda = 0.1$) and modality dropout ($p = 0.4$),
with Task~2 loss up-weighted at $w_{\mathrm{T2}} = 2$. Each ablation
row sets exactly one hyperparameter to its `off' value
($\lambda = 0$, $p = 0$, or $w_{\mathrm{T2}} = 1$) while keeping the
others at preferred values. All variants use same fixed seed and the same hyperparameters otherwise.}
\label{tab:ablation}
\begin{tabular}{lccccc}
\thickhline
\textbf{Configuration} & \textbf{$\lambda$} & \textbf{$p$} & \textbf{$w_{\mathrm{T2}}$} & \textbf{Task~1 AUROC [95\% CI]} & \textbf{Task~2 AUROC [95\% CI]} \\
\hline
Full (preferred)        & 0.10 & 0.40 & 2.0 & $0.797$ $[0.621, 0.948]$ & $0.755$ $[0.519, 0.945]$ \\
No contrastive          & 0.00 & 0.40 & 2.0 & $0.566$ $[0.344, 0.770]$ & $0.682$ $[0.427, 0.900]$ \\
No modality dropout     & 0.10 & 0.00 & 2.0 & $0.621$ $[0.406, 0.824]$ & $0.609$ $[0.346, 0.870]$ \\
Neither                 & 0.00 & 0.00 & 2.0 & $0.719$ $[0.512, 0.910]$ & $0.673$ $[0.398, 0.900]$ \\
No T2 up-weighting      & 0.10 & 0.40 & 1.0 & $0.668$ $[0.453, 0.863]$ & $0.673$ $[0.423, 0.889]$ \\
\thickhline
\end{tabular}
\end{table}

\subsection*{Comparison to tabular baselines}

To benchmark the multi-task fusion architecture, we trained three
baseline classifiers (logistic regression, XGBoost, TabMLP) under
two feature regimes: concatenated raw features with mean imputation,
and concatenated TabPFN-v2 embeddings (Methods). The TabPFN-embedding
baselines isolate the contribution of the architectural design from
the contribution of the upstream feature extractor. Test-set AUROCs
and bootstrap 95\% confidence intervals are shown in
Table~\ref{tab:baselines}.

On Task~1 (peak response), logistic regression on TabPFN embeddings
achieves AUROC $0.816$ ($[0.659, 0.957]$), narrowly exceeding the
multi-task model's $0.797$ point estimate; the two CIs overlap
heavily and the difference is not statistically distinguishable.
TabMLP on TabPFN embeddings is also competitive at $0.785$
($[0.623, 0.942]$). This indicates that, for peak prediction
alone, the per-modality TabPFN embeddings already carry most of
the predictive signal, and several classifiers on their
concatenation perform comparably to the full multi-task model. The
remaining Task~1 baselines fall below ours: XGBoost on raw
features ($0.781$), logistic regression and TabMLP on raw features
($0.699$ and $0.594$, both with lower CI below chance), and XGBoost
on TabPFN embeddings (degenerate at $0.500$).

On Task~2 (durability), the picture is sharply different. The
multi-task model achieves AUROC $0.755$ ($[0.519, 0.945]$),
substantially above all baselines on TabPFN embeddings: logistic
regression drops to $0.609$ ($[0.336, 0.857]$, lower CI below
chance), XGBoost to $0.364$ (below chance), and TabMLP collapses
to a degenerate constant predictor ($0.500$ across bootstrap
resamples, indicating training failure at this sample size). The
raw-feature baselines for Task~2 are stronger (logistic regression
$0.791$, TabMLP $0.777$) but their Task~1 performance is
correspondingly weak ($0.699$ and $0.594$, both with lower CI
below chance).

Read across both tasks, the multi-task fusion architecture is the
only method whose 95\% lower CI bound is above chance on both tasks
simultaneously. Every baseline either (i)~fails to clear chance on
Task~1 (raw-feature logistic regression and TabMLP), (ii)~fails to
clear chance on Task~2 (raw-feature XGBoost; TabPFN-embedding
logistic regression and XGBoost), or (iii)~degenerates entirely on
one of the two tasks (TabPFN-embedding XGBoost on Task~1, TabMLP
on Task~2). At this sample size, with this feature complexity and label structure, joint above-chance reliability across both biologically dissociated tasks requires the architectural design choices made here.

\begin{table}[!ht]
\centering
\caption{{\bf Test-set AUROC: comparison to baselines on raw features
and on TabPFN-v2 embeddings.} Bootstrap 95\% confidence intervals
from 1000 resamples. Baselines are fit on training subjects
($n_{\text{train}} = 94$ for Task~1, $n_{\text{train}} = 56$ for
Task~2) with default hyperparameters. The same seed for train/test
split is used throughout. ``Raw features'' baselines use
concatenated raw features from all four modalities with per-feature
training-set mean imputation; ``TabPFN emb.'' baselines use
concatenated 1,536-dimensional TabPFN-v2 embeddings (one per
modality, total 6,144 dimensions). Bold: best point estimate per
task; $\dagger$: lower 95\% CI bound below 0.5; $\ddagger$:
degenerate predictor (constant output, AUROC fixed at 0.5 across
all bootstrap resamples).}
\label{tab:baselines}
\small
\begin{tabular}{lcc}
\thickhline
\textbf{Method} & \textbf{Task~1 AUROC [95\% CI]} & \textbf{Task~2 AUROC [95\% CI]} \\
\hline
\multicolumn{3}{l}{\emph{Raw features (mean-imputed concatenation)}} \\
\hline
Logistic regression           & $0.699$ $[0.495, 0.867]^{\dagger}$ & $0.791$ $[0.555, 0.959]$ \\
XGBoost                       & $0.781$ $[0.599, 0.929]$           & $0.645$ $[0.382, 0.870]^{\dagger}$ \\
TabMLP                        & $0.594$ $[0.384, 0.801]^{\dagger}$ & $0.777$ $[0.526, 1.000]$ \\
\hline
\multicolumn{3}{l}{\emph{TabPFN-v2 embeddings (concatenation)}} \\
\hline
Logistic regression           & $\mathbf{0.816}$ $[0.659, 0.957]$  & $0.609$ $[0.336, 0.857]^{\dagger}$ \\
XGBoost                       & $0.500$ $[0.500, 0.500]^{\ddagger}$ & $0.364$ $[0.214, 0.500]^{\dagger,\ddagger}$ \\
TabMLP                        & $0.785$ $[0.623, 0.942]$           & $0.500$ $[0.500, 0.500]^{\ddagger}$ \\
\hline
Preferred (ours)              & $0.797$ $[0.621, 0.948]$           & $\mathbf{0.755}$ $[0.519, 0.945]$ \\
\thickhline
\end{tabular}
\end{table}

\section*{Discussion}

We addressed a challenging regime in precision vaccinology: predicting
both peak and long-term phases of the humoral booster response from
small, heterogeneous, partially missing multi-omic data. A multi-task
contrastive fusion architecture recovered predictive signal for both
phases (test AUROC $0.797$ for peak, $0.755$ for durability), with
both AUROCs significantly above chance under joint label permutation
($p = 0.002$ and $p = 0.045$ over $N = 1000$ permutations).

The two clinical endpoints are anti-correlated (Spearman
$r = -0.580$, Cohen's $\kappa = -0.520$), with most subjects sorting into
two response groups: large but fading, or modest but better retained.
The low peak / low durability quadrant is essentially empty ($n = 4$),
indicating that immune systems rarely fail on both axes simultaneously.
This dissociation is consistent with the canonical two compartment
biology of acute versus long-term humoral responses~\cite{Amanna2007,
Gillard2024} and is what motivates a multi-task framing rather than
single-task prediction of either phase alone.

Per-modality contribution analyses reveal that the trained model
routes the two tasks to different modalities. For peak prediction,
cytokine dominates: standalone AUROC $0.935$ exceeds the
all-modalities value ($0.888$), and removing cytokine causes the
largest LOO drop ($0.068$). Cell frequency is the weakest standalone
predictor ($0.783$) and its removal slightly \emph{improves}
ensemble performance, suggesting cell-frequency features are either
redundant with cytokine or actively noisy at this sample size. For
durability prediction, antibody alone matches the full-ensemble value ($0.735$)
while cytokine and cell drop sharply in isolation ($0.663$, $0.664$).
This task-specific routing is biologically coherent: peak response
at day~14 reflects acute B-cell activation and plasmablast expansion,
for which plasma cytokine signatures are direct mechanistic
markers~\cite{Pulendran2014, Querec2009}, whereas durability at day
120 depends on long-lived plasma cell maintenance, a process most
directly captured by the antibody titer trajectory
itself~\cite{Amanna2007}. The model recovers this underlying signature
without supervision on which modalities should matter for which task.

The architectural ablation directly tests whether multi-task
co-training contributes to peak prediction. At $w_{\mathrm{T2}} = 2$
(preferred), Task~1 AUROC is $0.797$; at $w_{\mathrm{T2}} = 1$, it
drops to $0.668$, a decline of $0.129$. This is direct evidence that
the Task~2 head acts as a regulariser on the shared backbone:
forcing the representation to also support durability prediction
prevents over-fitting to peak-specific idiosyncrasies, despite the
two tasks being biologically anti-correlated. The contrastive loss
and modality dropout components contribute on top of this base
multi-task benefit (Table~\ref{tab:ablation}); all three are
necessary to recover above-$0.79$ Task~1 AUROC.

The TabPFN-embedding baseline comparison clarifies the architectural design's distinctive contribution. On Task~1, the multi-task model and the strongest TabPFN-embedding
baselines achieve comparable AUROC (logistic regression $0.816$,
TabMLP $0.785$, multi-task $0.797$), with overlapping CIs. This is
consistent with the contribution analysis: peak is dominated by a
single highly informative modality that a linear classifier on
TabPFN embeddings can exploit directly. On Task~2, every baseline
either falls below chance or degenerates: durability requires
integration across modalities and a label structure (the dual-label
supervised contrastive loss) that no concatenation baseline can
access. The multi-task architecture is therefore the only method
tested that delivers above-chance performance on both tasks
simultaneously, and this joint reliability is its distinctive
contribution. The raw-feature baselines also face a structural disadvantage in handling missingness: with up to $38.6\%$ of subjects missing certain modalities, mean imputation fills entire modality blocks with training-set averages, producing biased input vectors for those subjects. The attention-based fusion in our architecture sidesteps this by excluding absent modalities from the softmax computation rather than imputing them.

The framework is naturally extensible to external validation cohorts
and to other vaccines for which similar multi-omic profiling exists,
allowing future tests of whether the peak-durability dissociation and
task-specific modality patterns reported here generalize beyond CMI-PB
and beyond pertussis.

\subsection*{Limitations}

This study has the following limitations.
First, the sample size is small ($n = 158$ subjects with a Task~1
label, $n = 96$ in the modeled Task~2 subset; $n_{\text{test}} = 32$
and $n = 21$ in the held-out test sets respectively), reflecting
the reality of longitudinal multi-omic human vaccine studies but
producing wide bootstrap confidence intervals on test AUROC, especially
for Task~2. We address this with permutation testing to confirm both
AUROCs are above chance, but a larger validation cohort will be needed
before clinical interpretation. Second, cross-cohort transportability has not been directly assessed: training on 2020-2022 and testing on 2023 would be a natural test, but is constrained by the absence of Task~2 labels in the 2023 cohort. External validation on independent multi-omic vaccine cohorts remains a goal for future work. Third, the per-modality
contribution analyses (LOO and KOO) describe the behavior of a
single trained model and should be interpreted as descriptions of
\emph{this} model's contributions rather than universal biological claims;
finer-grained, feature-level attribution within each modality is
deferred to future work. 

\section*{Data and code availability}

The CMI-PB data used in this study are publicly available through
the CMI-PB resource~\cite{CMIPB} at \url{https://www.cmi-pb.org}.
Code to reproduce the analyses is available at
\url{https://github.com/Divya1205/cmi-pb-multitask}.

\section*{Acknowledgments}

I thank the CMI-PB consortium for generating, harmonizing, and openly
sharing the longitudinal multi-omic pertussis booster vaccination
dataset that made this work possible, and Dr. Pramod Shinde for
helpful discussions and guidance on the dataset.

\nolinenumbers

\bibliographystyle{plos2025}
\bibliography{references}

\end{document}